\documentclass[conference]{IEEEtran}
\IEEEoverridecommandlockouts

\usepackage[square,sort,comma,numbers]{natbib}

\usepackage{booktabs}
\usepackage{multirow}

\usepackage{amsmath,amssymb,amsfonts}
\usepackage{algorithmic}
\usepackage{graphicx}
\usepackage{textcomp}
\usepackage{xcolor}
\usepackage{lipsum}
\usepackage[inline]{enumitem}

\usepackage{afterpage}

\usepackage{subcaption}

\usepackage{hyperref}
\hypersetup{colorlinks,allcolors=black}

\hypersetup{
       breaklinks=true,
       colorlinks=true,
       pdfusetitle=true,
    }

\def\BibTeX{{\rm B\kern-.05em{\sc i\kern-.025em b}\kern-.08em
    T\kern-.1667em\lower.7ex\hbox{E}\kern-.125emX}}
\begin{document}

\title{TrADe Re-ID -- Live Person Re-Identification using Tracking and Anomaly Detection
}

\author{Luigy Machaca$^1$, F. Oliver Sumari H.$^1$, Jose Huaman$^1$, Esteban Clua$^1$, Joris Guerin$^2$ \\
\textit{$^1$ Instituto de Computação, Universidade Federal Fluminense, Niterói, Brazil}\\
\textit{$^2$ Université Toulouse, LAAS-CNRS, ONERA, Toulouse, France} \\
luigyarcana@id.uff.br, fsumari@id.uff.br, jmhcruz@id.uff.br, esteban@ic.uff.br, jorisguerin.research@gmail.com
}

\maketitle

\begin{abstract}
Person Re-Identification (Re-ID) aims to search for a person of interest (query) in a network of cameras. In the classic Re-ID setting the query is sought in a gallery containing properly cropped images of entire bodies. Recently, the live Re-ID setting was introduced to represent the practical application context of Re-ID better. It consists in searching for the query in short videos, containing whole scene frames. The initial live Re-ID baseline used a pedestrian detector to build a large search gallery and a classic Re-ID model to find the query in the gallery. However, the galleries generated were too large and contained low-quality images, which decreased the live Re-ID performance. Here, we present a new live Re-ID approach called TrADe, to generate lower high-quality galleries. TrADe first uses a Tracking algorithm to identify sequences of images of the same individual in the gallery. Following, an Anomaly Detection model is used to select a single good representative of each tracklet. TrADe is validated on the live Re-ID version of the PRID-2011 dataset and shows significant improvements over the baseline. 
\end{abstract}

\begin{IEEEkeywords}
Person Re-ID, Tracking, Anomaly Detection
\end{IEEEkeywords}

\section{Introduction}
\label{sec:intro}

Video surveillance cameras are widely deployed in public places and are monitored continuously by human agents for public safety.
Although humans are able to analyze precisely any specific scene, they cannot monitor a large number of cameras simultaneously. 
Thus, the demand for automated pedestrian tracking systems is rapidly increasing~\cite{laufs2020security}. 
This paper deals with person Re-Identification (Re-ID), which consists in searching for a person-of-interest (query) in a network of non-overlapping cameras. 
The most common setting for Re-ID uses datasets of manually-cropped images containing only clean images representing full human bodies.
Then, the goal is to retrieve images from the search gallery that depict the query~\cite{standard_reID_survey2}. 
We refer to this setting as classic Re-ID.

Recent works have shown that classic Re-ID is not sufficient to implement useful real-world applications. 
In previous work, we introduced a new setting called live Re-ID, considering constraints related to implementing Re-ID for use during live operations~\cite{sumari2020towards}. 
Live Re-ID systems are composed of two main modules: the gallery generator, which extracts pedestrian bounding boxes, and the classic Re-ID module, which tries to identify the query from the cropped images in the gallery (Section~\ref{sec:Related_work_reID}).
Although most Re-ID research has focused on the latter, their experiments demonstrated that small errors in the gallery generation process can lead to poor live Re-ID results. 
In this work, we identified two properties of the object detectors used for gallery generation that limit the successful development of live Re-ID pipelines:
\begin{enumerate*}
    \item they often generate bounding boxes that do not represent entire human bodies, and
    \item they generate massive galleries, containing many correlated images of the same individuals, which impacts both the accuracy and execution time of the subsequent Re-ID module.
\end{enumerate*}

\begin{figure*}[t]
    \centering
    \begin{subfigure}[b]{0.32\textwidth}
        \centering
        \includegraphics[width=\textwidth]{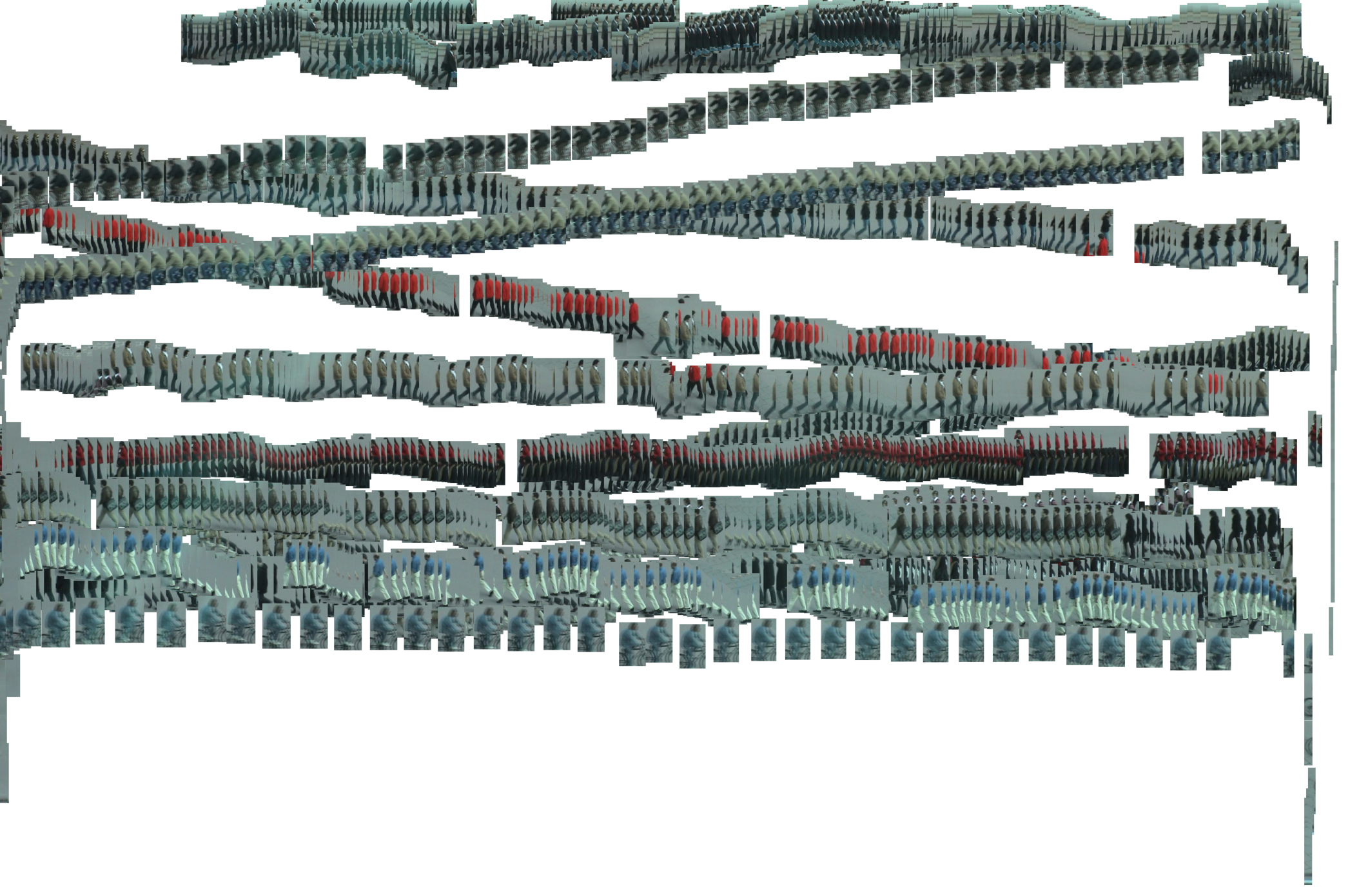}
        \caption{YOLOv3 gallery}
        \label{subfig:yolo_gallery}
    \end{subfigure}
    \hfill
    \begin{subfigure}[b]{0.32\textwidth}
        \centering
        \includegraphics[width=\textwidth]{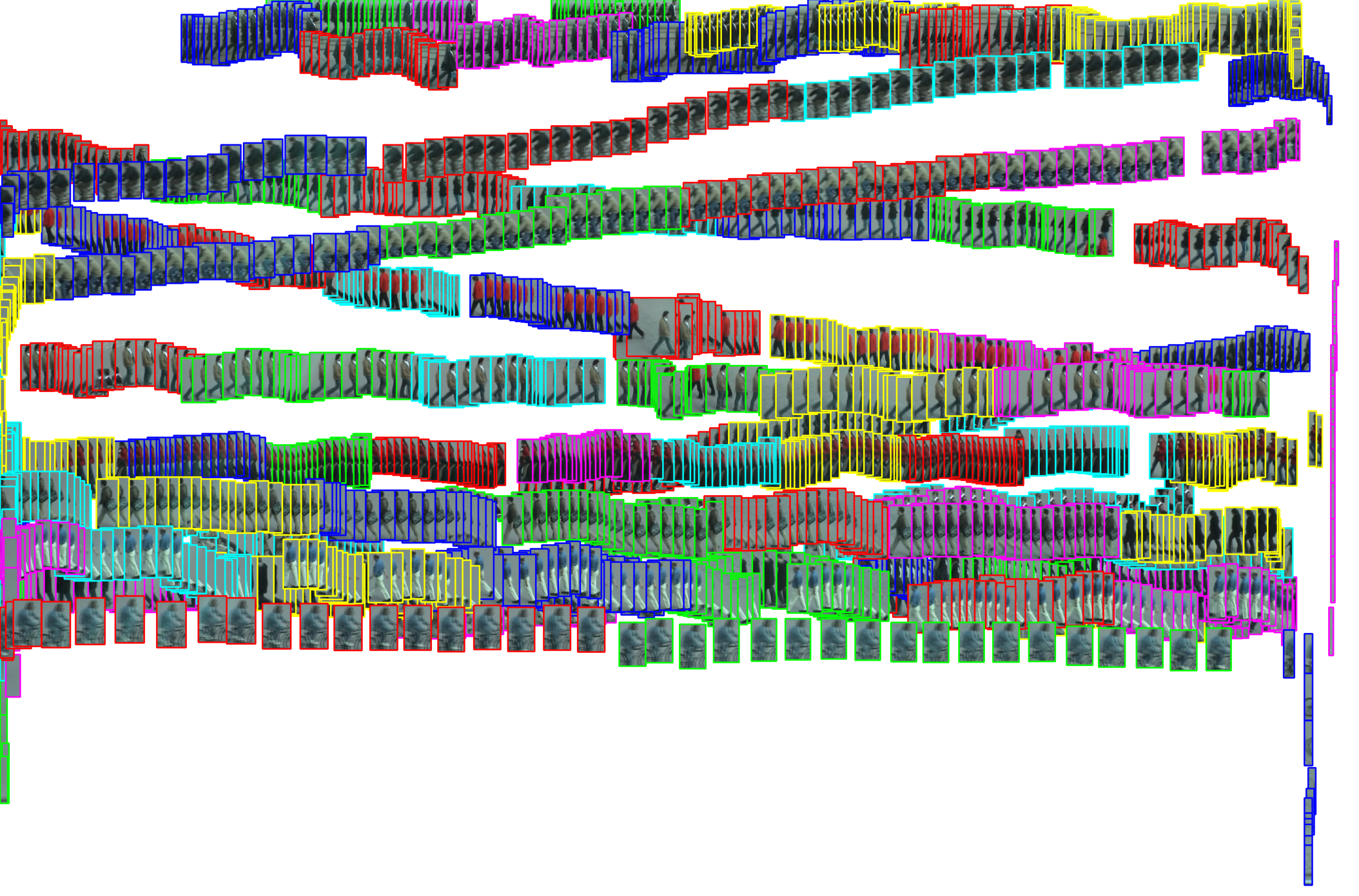}
        \caption{After Tracking}
        \label{subfig:after_tracking}
    \end{subfigure}
    \hfill
    \begin{subfigure}[b]{0.32\textwidth}
        \centering
        \includegraphics[width=\textwidth]{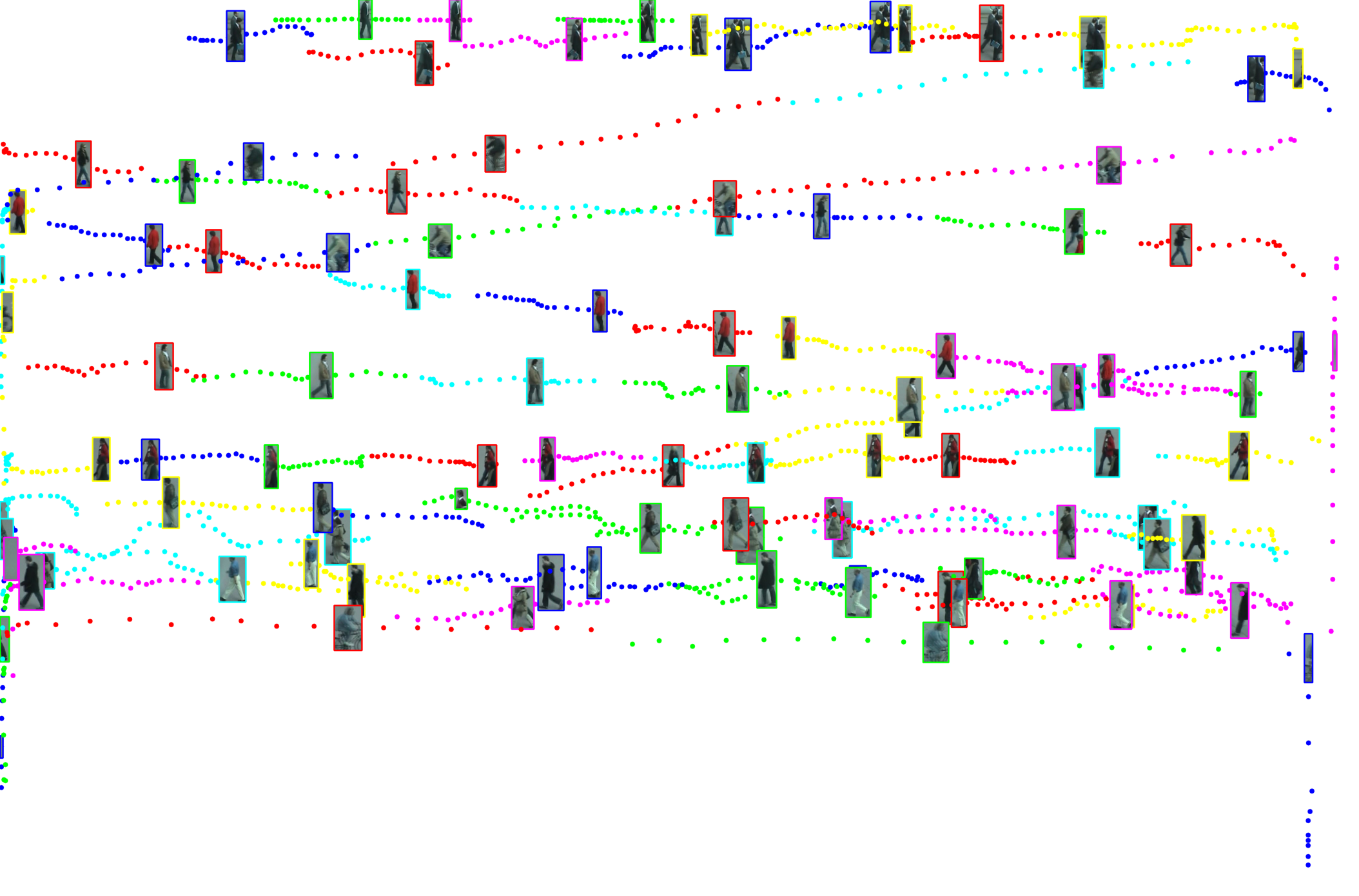}
        \caption{TrADe gallery}
        \label{subfig:trade_gallery}
    \end{subfigure}
    \caption{\textbf{Intuition behind TrADe live Re-ID}. This figure illustrates the TrADe steps on a $\sim$1.5 minute video from PRID2011. (a) Many pedestrian bounding boxes are extracted using YOLOv3 (gallery used in~\cite{sumari2020towards}, 1,945 images). (b) TrADe uses tracking to identify tracklets (adjacent boxes with the same color). (c) TrADe uses Anomaly Detection to select a single good bounding box to represent each tracklet. The new gallery is smaller ($\sim$100 images) and its images are better for Re-ID.}
    \label{fig:plot_tracklet_trajectory}
\end{figure*}

On the other hand, approaches from the field of video-based Re-ID have shown that using sequences of consecutive images of the same person can be valuable for Re-ID performance~\cite{zheng2016person}. Indeed, videos include much richer data than single images as we know that bounding boxes close to each other in space and time are likely to represent the same person. For example, in Figure~\ref{subfig:yolo_gallery} we can see that the standard gallery generation module generates a large number of bounding boxes, including poorly cropped ones near the edges. However, using Tracking, we can gain information and recover tracklets representing the same individuals (Figure~\ref{subfig:after_tracking}).

In this paper, we propose a novel live Re-ID approach to simultaneously reduce the size of the gallery and improve the quality of its images. This approach is called TrADe (gallery filtering using \textbf{Tr}acking and \textbf{A}nomaly \textbf{De}tection), and uses Object Tracking~\cite{gordon2018re3} to identify tracklets (consecutive bounding boxes corresponding to the same individual), and Anomaly Detection~\cite{perera2019learningDOC} to select a single good representative image of each tracklet (Figure~\ref{subfig:trade_gallery}). Figure~\ref{fig:live_ReID} clearly shows the steps involved in our TrADe pipeline.
We conduct experiments on the same live Re-ID dataset as \citet{sumari2020towards} and show that TrADe outperforms both their baseline approach and another baseline approach for gallery filtering. 
  

\section{Related work}
\label{sec:Related_work}

In recent years, techniques based on deep learning played an essential role in achieving good results at various computer vision tasks. This section presents the state-of-the-art about the different topics that are used in this work.

\subsection{Person Re-identification}
\label{sec:Related_work_reID}
Re-ID consists in retrieving instances of an individual (query), within a set of complex multimedia content (gallery). In the most popular Re-ID setting, which we call classic Re-ID, both the query and the items in the gallery are well-cropped images of a person's entire body~\cite{standard_reID_survey}. The person search setting uses galleries composed of whole scene images, to better represent the real-world application context of Re-ID~\cite{person_search_survey}. The open-set Re-ID setting extends classic Re-ID by adding the option that the query is not present in the gallery~\cite{open_set_survey}. Finally, in the video-based Re-ID setting, the query and gallery images are replaced by sequences of consecutive images from a video. Sequences contain clean full-body images representing the same individual~\citet{standard_reID_survey2}.

Recently, we introduced a new setting for real-world deployment of Re-ID, called live Re-ID~\cite{sumari2020towards, cruz2021benchmarking}, which combines elements from several of the above Re-ID settings. 
In practice, finding a query during live operations requires processing whole scene videos in near real-time. This way, 
The galleries for live Re-ID contain whole scene video frames. In addition, the probability that the query is present in a short video sequence from a given camera is low, which means that the live Re-ID setting is open-set. Finally, live Re-ID accounts for the fact that Re-ID predictions must be verified by human security agents, who can trigger actions. Hence, evaluation metrics were proposed to evaluate two objectives: high re-identification rate and low false alarm rate.

In this work, we aim to show that generating smaller galleries of higher quality images can substantially improve live Re-ID results, even without changing the classic Re-ID models. In our experiments, two classic Re-ID models are tested:
\begin{enumerate*}
    \item The Bag of Tricks (BoT) approach is based on several neural network training tricks rather than Re-ID architectural choices~\cite{luo2019bagBoT}.
    
    \item The SiamIDL approach uses a Siamese neural network architecture to predict whether two images represent the same person~\cite{ahmed2015improvedSiamIDL}.
\end{enumerate*}

\subsection{Object detection}
\label{sec:Related_work_od}
Object detection aims to locate object instances from predefined categories (e.g., pedestrians) in images. As it was already discussed extensively in a recent survey~\cite{liu2020deep}, we focus on presenting YOLO (You Only Look Once), the family of approaches used in this work.
YOLO architectures are sometimes referred to as unified detectors, or one-stage detectors, as they directly predict bounding boxes and class probabilities from full images, with a single forward pass over a Convolutional Neural Network. All versions of YOLO divide an image into grids and predict bounding box locations, each with class probabilities and associated confidence scores. For our experiments, we use YOLOv3~\cite{redmon2018yolov3}, which can detect small objects thanks to its multi-scale prediction capabilities.

\subsection{Object tracking} 
\label{sec:Related_work_tracking}
Object tracking aims to establish the location of a target object over the frames of a video sequence, starting from an initial bounding box.
In the literature, several surveys have proposed different classifications of object tracking approaches, e.g., single vs. multi-camera Tracking~\cite{iguernaissi2019people}, single vs. multi-object tracking~\cite{ondravsovivc2021siamese}, specific vs. generic object tracking~\cite{liu2020deep}.
In this work, we use an algorithm called Real-time, Recurrent, Regression-based tracker (Re3)~\cite{gordon2018re3}, which is an accurate, generic object tracker. Re3 uses convolutional layers to embed the object's appearance, recurrent layers to recall the appearance and motion of the object, and regression layers to output the object's location. Re3 requires a bounding box around tracked objects at the initial time step and produces bounding boxes in subsequent frames.

\subsection{Anomaly detection}
\label{sec:Related_work_occ}

Anomaly detection refers to the task of identifying data that significantly diverge from the patterns of expected data instances.
In recent years, deep learning approaches were used extensively to tackle anomaly detection, as shown in this recent survey~\cite{pang2021deep}. Most supervised deep anomaly detectors are composed of a feature extractor, to retrieve discriminative information that separates anomalies from regular instances, and a classifier. 
Among anomaly detection approaches, One-Class Classifiers (OCC) define a boundary around the native class (normal instances). At inference, an OCC generates a ``normality'' score that can be used to determine if it is an inlier or an outlier. In this work, we use DOC (Learning Deep Features for One-Class Classification)~\cite{perera2019learningDOC}. It uses a CNN for feature extraction, trained with two loss functions. 
The compactness loss fosters low intra-class distances by evaluating the closeness of the native class among the learned features.
The descriptiveness loss aims at finding large inter-class distances. Then, DOC uses a second neural network to produce the final classification score.

\begin{figure*}[t]
        \centering
        \includegraphics[width=0.95\textwidth]{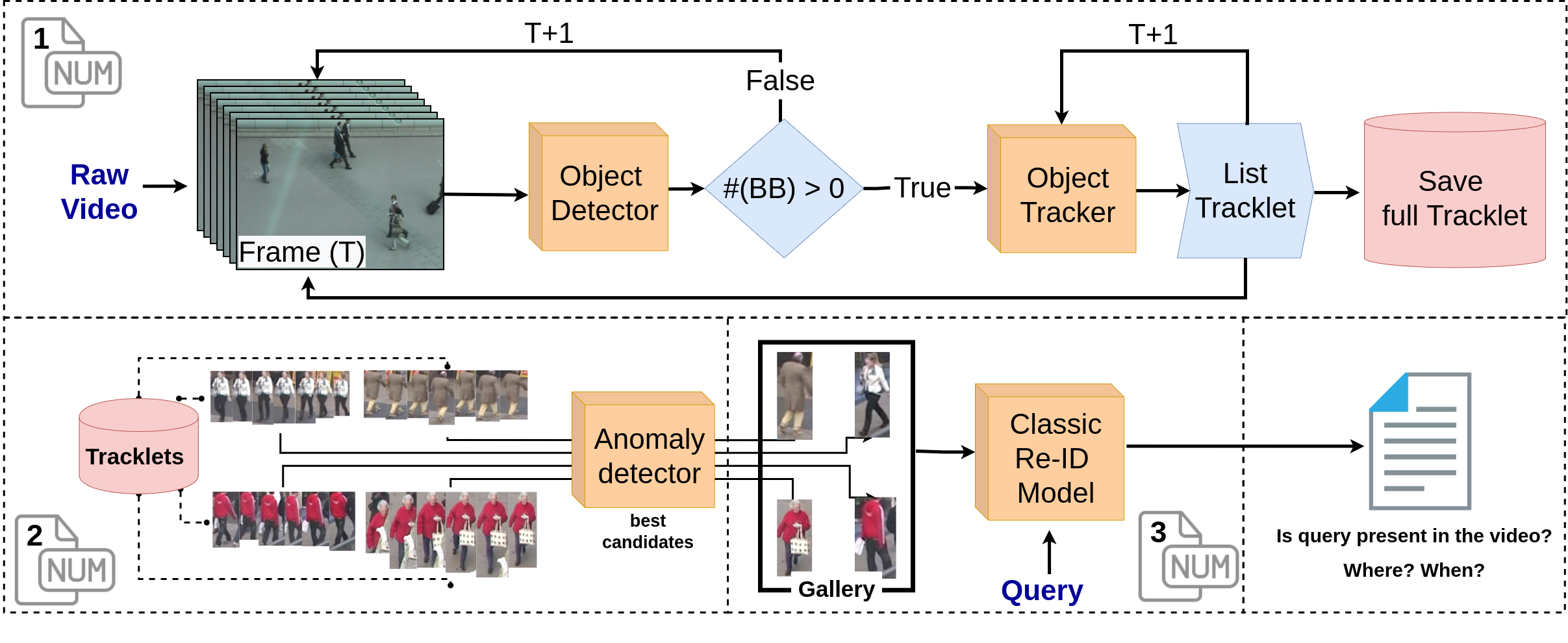}
        \caption{\textbf{Overview of TrADe live Re-ID}. This figure illustrates the general view of our proposal: (1) and (2) compose the gallery generation module, and (3) is the classic re-identification module of live Re-ID. The whole pipeline uses a raw video and a query image as inputs and returns the list of most similar detections and their corresponding scores.
        }
        \label{fig:live_ReID}
\end{figure*}

\section{TrADe Re-ID methodology}
\label{sec:Methodology}

This section presents the different components of TrADe and further information about the implemented pipeline.

\subsection{Overview of the approach}

A live Re-ID pipeline receives as input a short video sequence and a query image. It returns whether the query is present in the video, as well as information regarding where and when it appears. The baseline live Re-ID pipeline proposed in \cite{sumari2020towards} uses a YOLOv3 object detector to locate pedestrians in every frame of the video. The obtained bounding boxes make up the search gallery in which the query is sought using a classic Re-ID model (SiamIDL). The issue with this approach is that it generates very large galleries, containing some very bad images due to errors of the object detector. 

Here, we introduce TrADe (Figure~\ref{fig:live_ReID}), which is able to reduce the gallery size and improve the quality of its images. It relies on using a Tracking algorithm to identify bounding boxes representing the same individual in consecutive frames (tracklets). Then, Anomaly Detection is used to select a single good image to represent each tracklet. Lastly, a classic Re-ID model is used to compute the similarity scores between the query and the images of the gallery.

\subsection{Generating the tracklets}\label{sec:tracklet_gen}

The first step is to generate short sequences of consecutive images representing the same individual. To do this, the first frame of the search video is processed by the object detector to generate initial bounding boxes. Then, the detections are used to initialize the object tracker, which runs in the following frames to generate a tracklet. 

One of the main issues with modern tracking algorithms is the label-switching problem. It happens when people cross, or when one goes out of the frame and another enters a few frames later at a nearby location. This can lead to very long tracklets, containing different persons, which is an undesirable property for TrADe. Indeed, as TrADe only selects a single bounding box to represent an entire tracklet, if several persons appear in the same tracklet, some might not be represented in the final gallery. Hence, we force TrADe to generate small tracklets by fixing their maximum length $N$ (user-defined parameter). In practice, whenever a tracklet contains $N$ frames, it is stopped. Then, the object detector is run every $N$ frames to initialize new tracklets. The influence of the parameter $N$ on live Re-ID results is evaluated in our experiments.

\subsection{Selecting a single image to represent a tracklet}

Once short tracklets have been generated, we want to select a single good image for each tracklet to enter the search gallery. A good image is defined as a properly cropped image containing the entire body of a single human being. In other words, we want to generate galleries that contain images belonging to the domain of classic Re-ID training datasets.

To select the gallery images, we use an anomaly detection approach called DOC~\cite{perera2019learningDOC}. It is a one-class classifier trained to distinguish good images for Re-ID from bad ones. For each image, DOC produces a score representing its fitness for Re-ID. This score is then used to select the best image of the tracklet, i.e., the one with the highest score.

To be able to train the DOC classifier, we need sufficient example images of the target class (good images for classic Re-ID), as well as counter-examples, representing bad images for classic Re-ID. To build this DOC training dataset, we use a subset of the CUHK03~\cite{li2014deepreidCUHK03} Re-ID dataset as target examples, and the non-target examples are collected from the VOC2012 dataset~\cite{everingham2010pascal} (images not representing humans).

\subsection{Practical implementation choices}

Here, we present the implementation details to reproduce our results. The complete code is available on github\footnote{\url{https://github.com/luigy-mach/TrADe}}.

\paragraph{Pedestrian detection}
We use the pre-trained YOLOv3 with Darknet-53 backbone proposed in TensorFlow, which was trained on VOC 2012~\cite{everingham2010pascal}. We only use the ``person'' output class and bounding boxes with a score below 0.5 are rejected.

\paragraph{Pedestrian tracking}
For tracking, we used the pre-trained Re3 model from the official repository\footnote{\url{https://github.com/danielgordon10/re3-tensorflow}}. 

\paragraph{Anomaly detection}
DOC was trained using images from CUHK03~\cite{li2014deepreidCUHK03} as the target class, and images from VOC 2012 as outliers. The backbone of the feature extractor is an InceptionResNetV2~\cite{szegedy2017inception} pre-trained on ImageNet.

\paragraph{Classic Re-ID}
For SiamIDL~\cite{ahmed2015improvedSiamIDL}, we use the same implementation as \cite{sumari2020towards}, trained on CUHK03.
For BoT~\cite{luo2019bagBoT}, we use a ResNet-50 backbone~\cite{he_deep_2015} pre-trained on ImageNet and fine-tuned on Market-1501~\cite{zheng2015scalableMARKET}.

\section{Experimental evaluation}

In this section, we describe the details of the experimental evaluations conducted in this work.


\subsection{Dataset}
\label{sec:dataset_PRID}
To the best of our knowledge, today there is only one public dataset that can be used to evaluate complete live Re-ID pipelines, which was introduced in~\cite{sumari2020towards}. It is a modified version of the PRID-2011 dataset~\cite{prid2011}, based on the raw video footage and the original annotations that were used to create the official version of PRID-2011. The PRID-2011 videos were collected from two non-overlapping cameras located in Graz, Austria. It contains 385 identities for the first camera and 749 for the second, with 200 shared identities across both cameras. The modified PRID (live-PRID) dataset contains several two minutes videos (63), and for each short video, it has a ground truth file associated with information about each individual it contains. For evaluation, we consider 73 queries in total.

To evaluate TrADe, we apply the same evaluation methodology as~\citet{sumari2020towards}. We select ten videos of two minutes from each camera. Between each pair of videos, we select the persons who appear in both cameras. Approximately the first four query images for each video were selected and exchanged between each video by ensuring that each query appears at least in one frame. 
Using the notations from \citet{sumari2020towards}, we use the following parameters for our live Re-ID pipeline: 
\begin{itemize}
    \item The number of frames for video splitting ($\tau$) is set to $1000$, the best value from the experiments in~\cite{sumari2020towards}.
    \item For $\beta$, the threshold on Re-ID scores for generating an alert to the monitoring agent, we use values between 0 and 1 with a step size of $0.02$.
    \item The number of candidates shown to the monitoring agent ($\eta$) is set to $20$, which was also the best value in \cite{sumari2020towards}.
\end{itemize}  




\subsection{Evaluation metrics}
\label{sec:evaluation_metrics}

To assess the performance of TrADe Re-ID on the live-PRID dataset, we use the evaluation metrics introduced in~\cite{sumari2020towards}:

\begin{itemize}
    \item the Finding Rate (FR) is the proportion of short videos such that the query was present and presented to the monitoring agent. A low FR occurs when the query is missed frequently.
    
    \item the True Validation Rate (TVR) is the proportion of alerts raised to the agent such that the query was among the presented candidates. A low TVR occurs when thez monitoring agent was frequently unjustified disturbed . 
\end{itemize} 

To better present the results, we use the following two metrics to ease the interpretation of TrADe Re-ID results:
\begin{itemize}
    \item Similarly to the mean Average Precision (mAP) for standard object detection approaches, we compute the area under the \textit{TVR vs FR} curve and call it mAP by analogy. This allows to present results that are independent of the threshold $\beta$.
    \item Similarly to the F-score computation for precision and recall, we compute the $F_{1}$ score for FR and TVR as their harmonic mean. 
    However, each value of the threshold $\beta$ involves a different value of $F_1$. To address this problem, we use the optimal configuration for $F_1$ and call it $F^*_1$ (see~\cite{icmla_joris}). In other words, it corresponds to the highest value of $F_1$ among all possible values of $\beta$. An $F^*_1$ of 1 means that there exists a Re-ID threshold $\beta$ such that the live Re-ID pipeline works perfectly. When single values of FR and TVR are reported, they are the ones corresponding to the optimal $F_1$ threshold.
\end{itemize} 

\subsection{Comparison with other approaches}
\label{sec:comparison_with_other_approches}
Our TrADe Re-ID approach for live Re-ID is compared against the baseline presented in~\cite{sumari2020towards}, which corresponds to the limit case when the maximum length of the tracklet ($N$) is set to one. To evaluate if the benefits of TrADe are only due to the reduced gallery size, we also compare TrADe against a simpler approach for gallery size reduction, which we call \emph{Skip}. It consists in simply running the YOLOv3 object detector once every $N$ frames, where $N$ is the maximum tracklet size for TrADe. This simple approach generates galleries of the same size as TrADe and allows us to evaluate the impact of the anomaly detection component of TrADe.

\section{Results}
\label{sec:Results}

We now present and discuss the results obtained to validate the effectiveness of our new strategy for live Re-ID.

\subsection{Performance of TrADe}
\label{sec:results_trade}

The results obtained with different approaches are reported in Table~\ref{table:all_results}. The results presented are for $N=20$, which appears to produce a good trade-off between gallery size reduction and loss of information.  

\begin{table}[t]
\centering
\caption{\textbf{live-PRID results}. Results obtained with different live Re-ID approaches (including TrADe) on the live-PRID dataset. These results are for $N=20$.}
\begin{tabular}{@{}llllll@{}}
\toprule 
& & FR & TVR & $F^{*}_{1}$ & mAP \\ \midrule
\multirow{3}{*}{SiamIDL} & Baseline~\cite{sumari2020towards} & 0.544 & 0.196 & 0.289 & 0.104 \\
& Skip & 0.792 & \textbf{0.500} & 0.422 & 0.258 \\
& TrADe & \textbf{0.823} & \textbf{0.500} & \textbf{0.439} & \textbf{0.279} \\ \midrule
\multirow{3}{*}{BoT} & Baseline~\cite{sumari2020towards} & 0.506 & 0.188 & 0.268 & 0.095 \\
& Skip & 0.835 & \textbf{0.387} & 0.463 & 0.302 \\
& TrADe & \textbf{0.886} & 0.372 & \textbf{0.481} & \textbf{0.317} \\ \bottomrule
\end{tabular}
\label{table:all_results}
\end{table}

We can see that for both classic Re-ID approaches (SiamIDL and BoT), it is generally a good idea to reduce the gallery size. Indeed both the simple approach (Skip) and TrADe perform significantly better than the baseline from \citet{sumari2020towards}. This means that galleries generated by simply using the object detector on every frame are too large to be processed correctly by the classic Re-ID models and generate noise.

We can also see that TrADe performs almost always better than Skip. This means that using the Anomaly Detection module for selecting good images to represent tracklets is a good idea for live Re-ID. Overall, these results confirm that TrADe is a promising approach to address the live Re-ID problem, leading to significant improvements over the current state-of-the-art baseline proposed in~\cite{sumari2020towards}.

\subsection{Influence of the maximum tracklet size $N$} \label{sec:results_tracklet_size}

\begin{figure}[t]
    \centering
\begin{subfigure}[b]{0.495\textwidth}
    \centering
    \includegraphics[width=\textwidth]{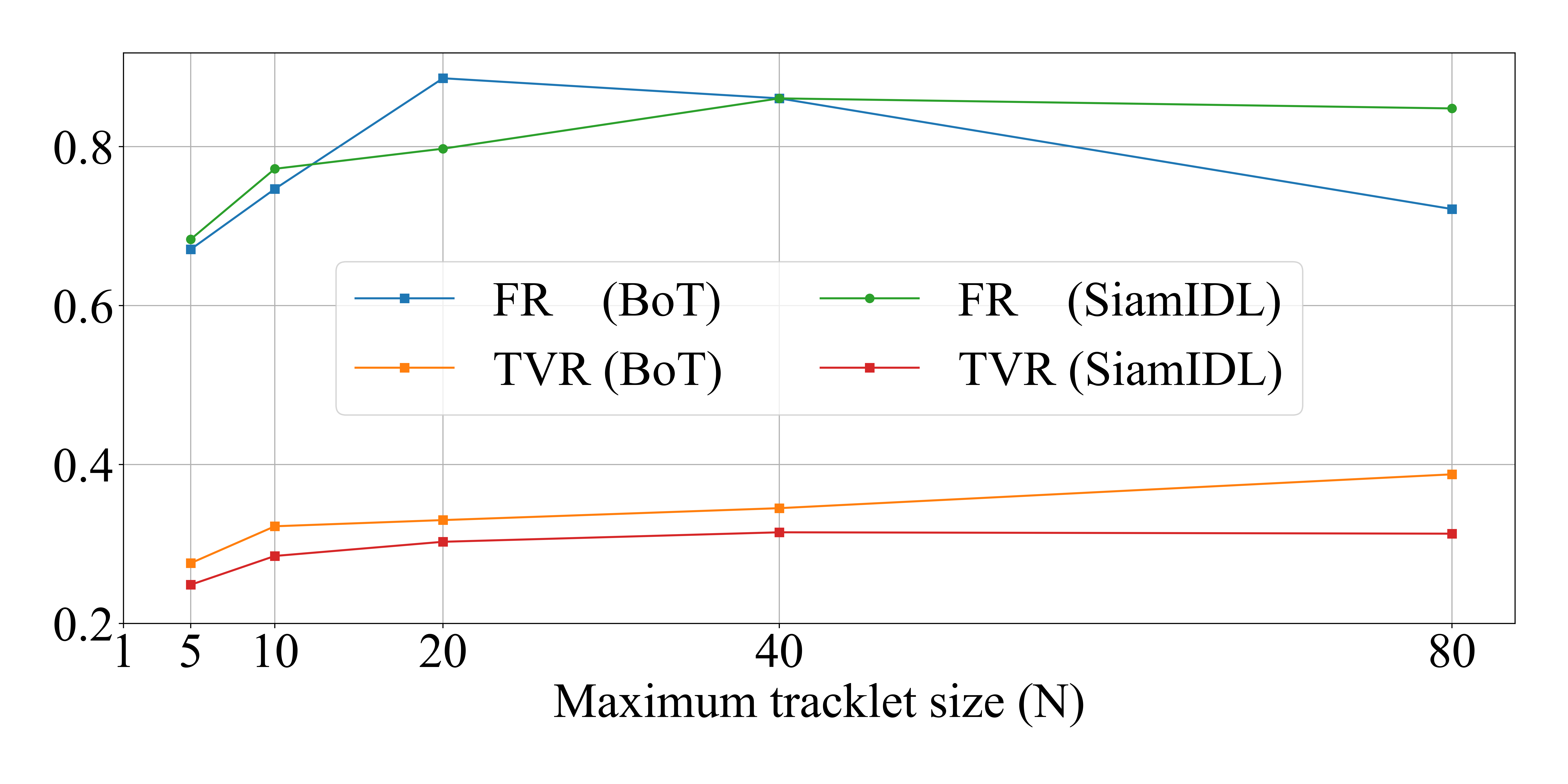}
    \caption{Finding Rate(FR) and True Validation Rate(TVR) through tracklet size (N)}
    \label{fig:influenceN_f1_map}
\end{subfigure}

\begin{subfigure}[b]{0.495\textwidth}
    \centering
    \includegraphics[width=\textwidth]{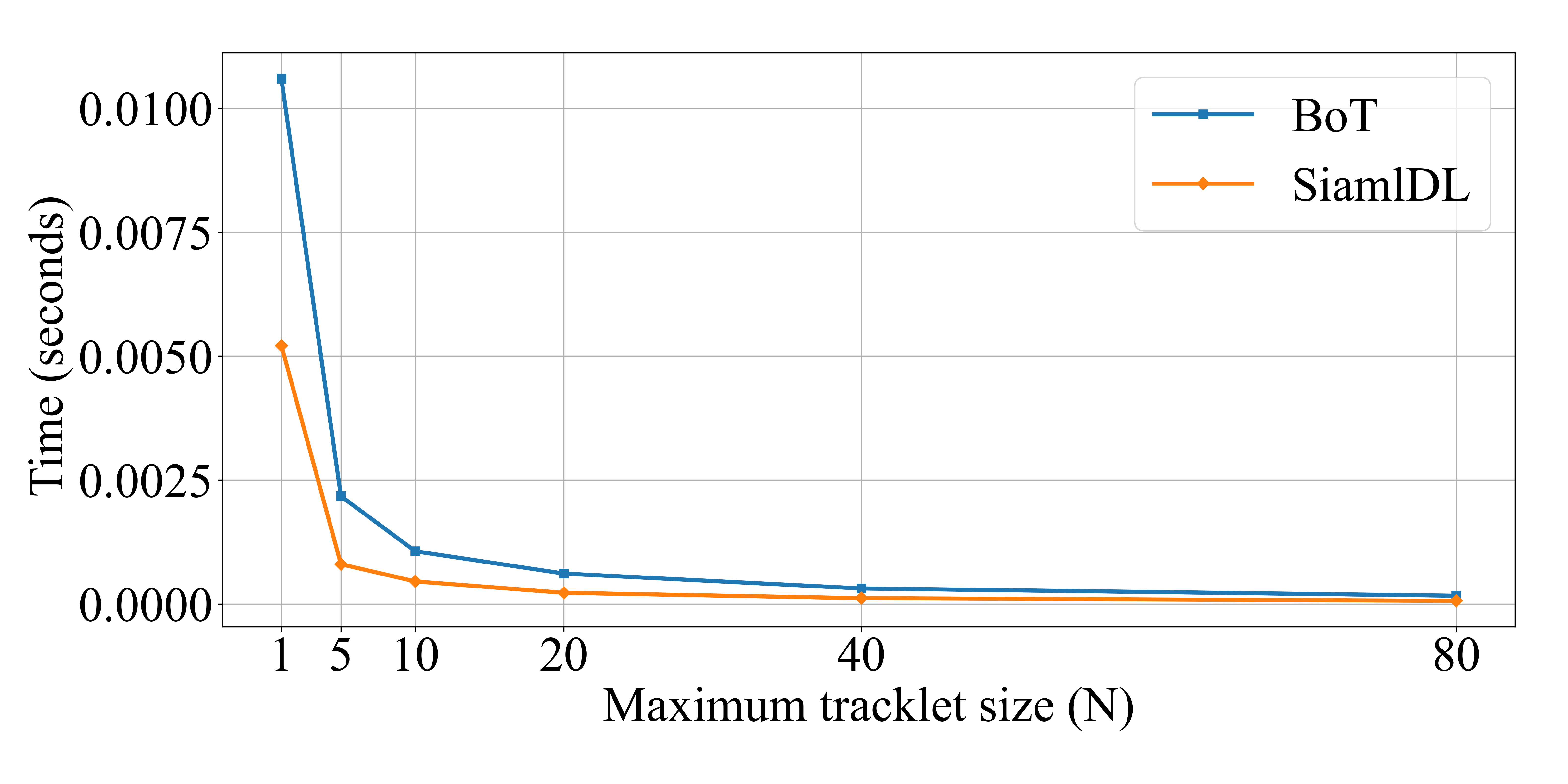}
    \caption{Average processing time for the classic Re-ID module}
    \label{fig:influenceN_time}
\end{subfigure}
\caption{\textbf{Influence of maximum tracklet length}. These graphs show the values taken by several important metrics (mAP, $F_1^*$, classic Re-ID execution time) for different values of the hyperparameter $N$.
}
    \label{fig:influenceN}
\end{figure}

Here, we discuss the influence of the hyperparameter $N$, which is the maximum length of a tracklet on the live Re-ID results obtained with TrADe. We test different values of $N$ on a log-scale: $\{1, 5, 10, 20, 40, 80\}$, and the curves representing the evolution of $F_1^*$, mAP, and the time required to run classic Re-ID on the generated gallery are shown in Figure~\ref{fig:influenceN}.

Figure~\ref{fig:influenceN_f1_map} shows that increasing $N$ helps to improve the TVR (True Validation Rate). This makes sense as larger values of $N$ lead to smaller galleries, containing less misleading images, which in turn generate fewer false alarms. On the other hand, we can see that this pattern is less clear for the FR (Finding Rate), which starts decreasing for BoT when $N$ is above $20$. This also makes sense because when we allow for very long tracklets, some persons will not appear in the final gallery due to the label switching problem (see Section~\ref{sec:tracklet_gen}). For these reasons, we used a value of $N=20$ in the experiments of Section~\ref{sec:results_trade}.

The second result is that the classic Re-ID processing time appears to decrease drastically as we increase the size of the tracklet $N$ (Figure~\ref{fig:influenceN_time}). This makes intuitive sense, as when $N$ increases, the size of the gallery decreases, and by extension the classic Re-ID module needs to process fewer images. We also note that the time decreasing effect is less pronounced when $N$ exceeds 20. This is because most images in the gallery can already fit in a single batch for GPU processing at that point. 

\section{Conclusions and future works}
\label{sec:Conclusions_and_future_works}

In this work, we addressed the live Re-ID problem, which uses raw videos as search galleries instead of manually cropped full-body images. A first baseline approach for live Re-ID was proposed in~\cite{sumari2020towards}, using object detection to generate a search gallery, and classic Re-ID to find the query in the gallery. A major issue with this baseline is the fact that the galleries obtained are too large, and contain outlier images, which do not represent human bodies. In this work, we propose to use a tracking algorithm to identify when successive bounding boxes are of the same individual and group them as tracklets. Following, an anomaly detection model is used to select the ``most normal'' image of each tracklet. This approach is called TrADe and generates lower galleries than the baseline, with fewer outliers. Our experimental results confirm that TrADe performs much better than the baseline, which is a huge step toward building better Re-ID applications. 

We present two ideas that could be explored in future works. First, our pipeline uses several deep neural networks that were all pre-trained on ImageNet (initial layers). Hence, we could speed up the pipeline considerably by building a single architecture, sharing these generic first layers among the four modules based on deep learning. A second promising idea would be to not only consider a single image per tracklet but rather several good quality images that are complementary, i.e., representing different poses of a person. The good results obtained recently on the video-based Re-ID setting~\cite{shu2021diverse} suggest that this could have a positive impact on live Re-ID results.
\bibliographystyle{IEEEtranN.bst}
\bibliography{unorganized.bib} 

\end{document}